\documentclass[letterpaper, 10 pt, conference]{ieeeconf}  

\IEEEoverridecommandlockouts                              

\usepackage{cite}
\usepackage{graphics} 
\usepackage{epsfig} 
\usepackage{epstopdf}
\graphicspath {{./image/}}
\usepackage{mathptmx} 
\usepackage{times} 
\usepackage{amsmath} 
\usepackage{amssymb}  
\usepackage{gensymb}
\usepackage{multirow}
\usepackage{picinpar}
\usepackage{multicol}
\usepackage{makecell}
\usepackage{booktabs}
\usepackage{stfloats}
\usepackage{caption}
\usepackage{graphicx}
\usepackage{subfigure}
\usepackage{wasysym}
\usepackage{textcomp}
\usepackage{siunitx}
\let\labelindent\relax
\usepackage{enumitem}
\usepackage{comment}
\usepackage[colorlinks,linkcolor=red,anchorcolor=blue,citecolor=green]{hyperref}
\def\BibTeX{{\rm B\kern-.05em{\sc i\kern-.025em b}\kern-.08em
    T\kern-.1667em\lower.7ex\hbox{E}\kern-.125emX}}

\usepackage{xcolor}

\title{\LARGE \bf
RadarLoc: Learning to Relocalize in FMCW Radar
}

\author{Wei Wang$^{1}$, Pedro P.\ B.\ de Gusm\~{a}o$^{1}$, Bo Yang$^{2}$, Andrew Markham$^{1}$, and Niki Trigoni$^{1}$
\thanks{$^{1}$The authors are with the Department of Computer Science, University of Oxford, OX1 3QD, United Kingdom.
        {\tt\small \{firstname.lastname\}@cs.ox.ac.uk}}%
\thanks{$^{2}$ Bo Yang is with the Department of Computing, The Hong Kong Polytechnic University, HKSAR. {\tt\small bo.yang@polyu.edu.hk} }%
}

\begin{document}

\maketitle
\thispagestyle{empty}
\pagestyle{empty}

\begin{abstract}

Relocalization is a fundamental task in the field of robotics and computer vision. There is considerable work in the field of deep camera relocalization, which directly estimates poses from raw images. However, learning-based methods have not yet been applied to the radar sensory data. In this work, we investigate how to exploit deep learning to predict global poses from Emerging Frequency-Modulated Continuous Wave (FMCW) radar scans. Specifically, we propose a novel end-to-end neural network with self-attention, termed RadarLoc, which is able to estimate 6-DoF global poses directly. We also propose to improve the localization performance by utilizing geometric constraints between radar scans. We validate our approach on the recently released challenging outdoor dataset Oxford Radar RobotCar. Comprehensive experiments demonstrate that the proposed method outperforms radar-based localization and deep camera relocalization methods by a significant margin.

\end{abstract}

\section{INTRODUCTION}

Relocalization is a fundamental problem in robotics and computer vision. A robot has to localize itself when moving in urban or indoor environments to achieve competent autonomy. Several existing solutions employ Global Navigation Satellite System (GNSS) to perform localization. However, GNSS is not always available such as in indoor environments and the accuracy of GNSS cannot be guaranteed in urban environments with high-rising buildings since they can block GNSS signals. There is a significant body of knowledge in visual localization, as it has been studied for decades. Conventional geometry-based visual localization systems mainly utilize handcrafted features and descriptors, which are typically sensitive to illumination variation, dynamic objects and viewpoint change~\cite{Huang2019PriorEnvironments}. Recently, learning-based visual localization methods such as PoseNet and variants~\cite{Kendall2015,Kendall2016ModellingRelocalization,Kendall2017,Brahmbhatt2018} have been proposed to solve these challenges, which leverage either a single image or a sequence of images to predict 6-Degree-of-Freedom (6-DoF) poses directly. Unlike retrieval-based learning approaches e.g. CamNet~\cite{Ding2019CamNetRe-Localization}, RelocNet~\cite{Balntas2018RelocNetNets} and Camera Relocalization CNN~\cite{Laskar2017CameraNetwork}, location-related information of these deep learning methods is implicitly encoded within the parameters of these deep neural networks, and therefore these methods require agents that have previously traversed the same environment. However, vision sensors inherently suffer from several drawbacks which restrict their ability to be used in scenarios where reliability is highly desirable, such as self-driving cars. Visual inputs are easily impacted by ambient environmental conditions e.g. sunshine, rain, fog; and further by their narrow Field-of-View (FoV).
    
	Emerging Frequency-Modulated Continuous Wave (FMCW) radar sensors can effectively solve many of the shortcomings of cameras. They can provide a $360\degree$ view of the scene and range objects hundreds of meters away. Meanwhile, they can function reliably in unstructured environments in different conditions e.g. snow, darkness, fog, smoke, direct sunlight~\cite{Barnes2020TheDataset} without impact. These characteristics of radar make it suitable for robot localization, especially for autonomous agents which operate in large-scale urban scenes. Inspired by the aforementioned deep pose regression methods that use images, the aim of this work is to investigate and provide a robust radar localization system, allowing robots to relocalize themselves under previously visited scenes. 

\begin{figure}
  \centering
  \includegraphics[width=\columnwidth]{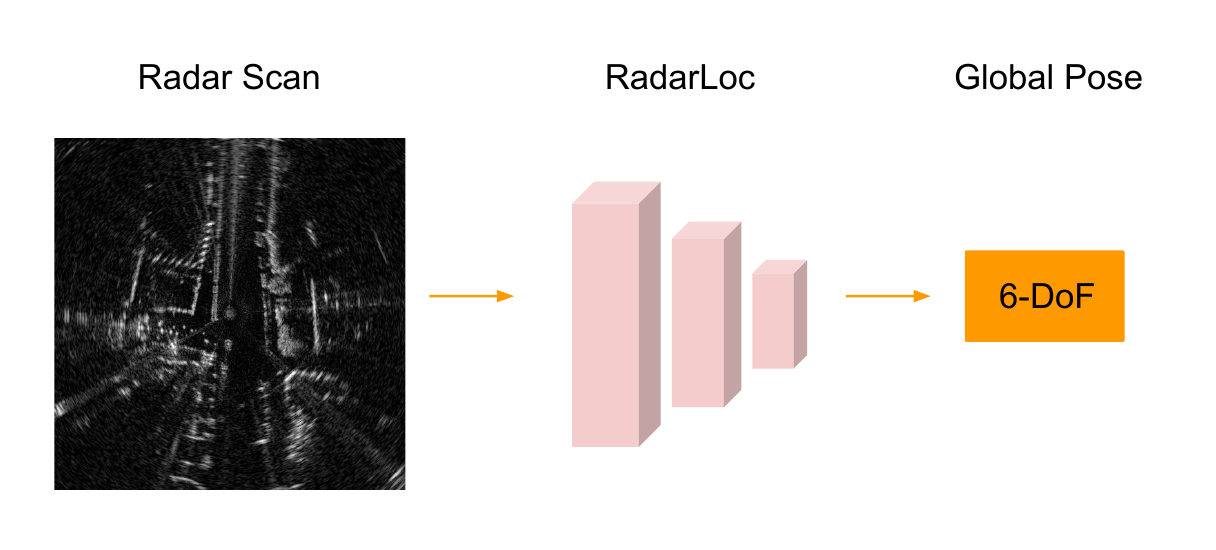}
  \caption{System overview of the proposed RadarLoc relocalization framework. A raw FMCW radar scan is first transformed into a Cartesian Radar Image. The radar image is then fed to RadarLoc, which directly estimates the 6-DoF pose in an end-to-end manner.}
  \label{fig:system_overview}
\end{figure}

    Specifically, we propose a novel geometry-aware neural network architecture, termed RadarLoc, which can estimate the 6-DoF pose using a single radar scan. The proposed self-attention module of a nested encoder-decoder architecture further improves the localization performance. During the training phase, RadarLoc takes as input a sequence of radar scans, and predicts poses optimized and constrained by both absolute and relative (geometric) pose losses. At inference time the 6-DoF pose is regressed from a single input scan. Fig.~\ref{fig:system_overview} illustrates the overview of the proposed fully differentiable relocalization system. 

    Our contributions are summarized as follows: 
We demonstrate that radar scans can be employed to estimate absolute 6-DoF poses in an end-to-end fashion. We further refine pose estimations by leveraging geometric constraints between radar pairs as one component of the loss function. Comprehensive experiments and ablation study have been done to demonstrate the effectiveness of RadarLoc, which outperforms state-of-the-art radar-based localization, DNN-based camera relocalization methods by a significant margin.
	
    
\section{RELATED WORK} \label{related_work}





\subsection{Deep Camera Localization}
Apart from problems of computation and storage, traditional visual localization in dynamic environments is still very difficult because of foreground outliers and appearance variations \cite{Huang2019PriorEnvironments}. For tackling these problems, recent works propose DNN-based methods to estimate 6-DoF poses directly. Single or sequential images are fed into a neural network model which comprises a feature extractor and a pose regressor for estimating absolute poses in an end-to-end manner. PoseNet~\cite{Kendall2015} is the first to demonstrate that 6-DoF camera poses can be directly predicted by a neural network. Following variations \cite{Kendall2016ModellingRelocalization,Kendall2017} improve the performance of PoseNet by introducing a geometric loss and modelling the uncertainty of poses with Bayesian Neural Network. Walch \textit{et al.}~\cite{Walch2017} proposed to utilize LSTM for structural feature correlation to improve the performance. 
Although these approaches are promising, they are still limited to the disadvantages of visual sensors. Our work extends this line of research by leveraging FMCW scanning radar to perform deep global localization. 

\subsection{Radar Geometry}
A $360\degree$ FMCW radar continuously scans the surrounding environment with a total of $M$ azimuth angles. The radar emits a beam and collapses the return signal for each azimuth angle~\cite{Hong2020RadarSLAM:Weathers}. The raw scan of the FMCW radar is a polar image, which can be transformed into a Cartesian image. Formally, given a point $(a, b)$ where $a$ is the azimuth and $b$ is range on a raw polar image, the range angle $\theta$ in the corresponding Cartesian coordinate is:
\begin{equation}
	\theta\ =\ 2\pi\cdot a\ /\ M
\end{equation}
 Thus, the corresponding coordinate $\mathbf{Z}$ in the Cartesian image can be calculated as:

\begin{equation}
	\mathbf{Z}\ =\ \begin{bmatrix}
                            \alpha \cdot cos\theta \cdot b\\
                            \alpha \cdot sin\theta \cdot b
                            \end{bmatrix}
\end{equation}
where $\alpha$ is a scaling factor between the image pixel space and the world metric space. Cartesian representation of the radar scan is visually comprehensible, and is better for neural networks to learn and optimize than the raw polar representation.

\subsection{Radar Odometry}
\begin{figure}
  \centering
  \includegraphics[width=\columnwidth]{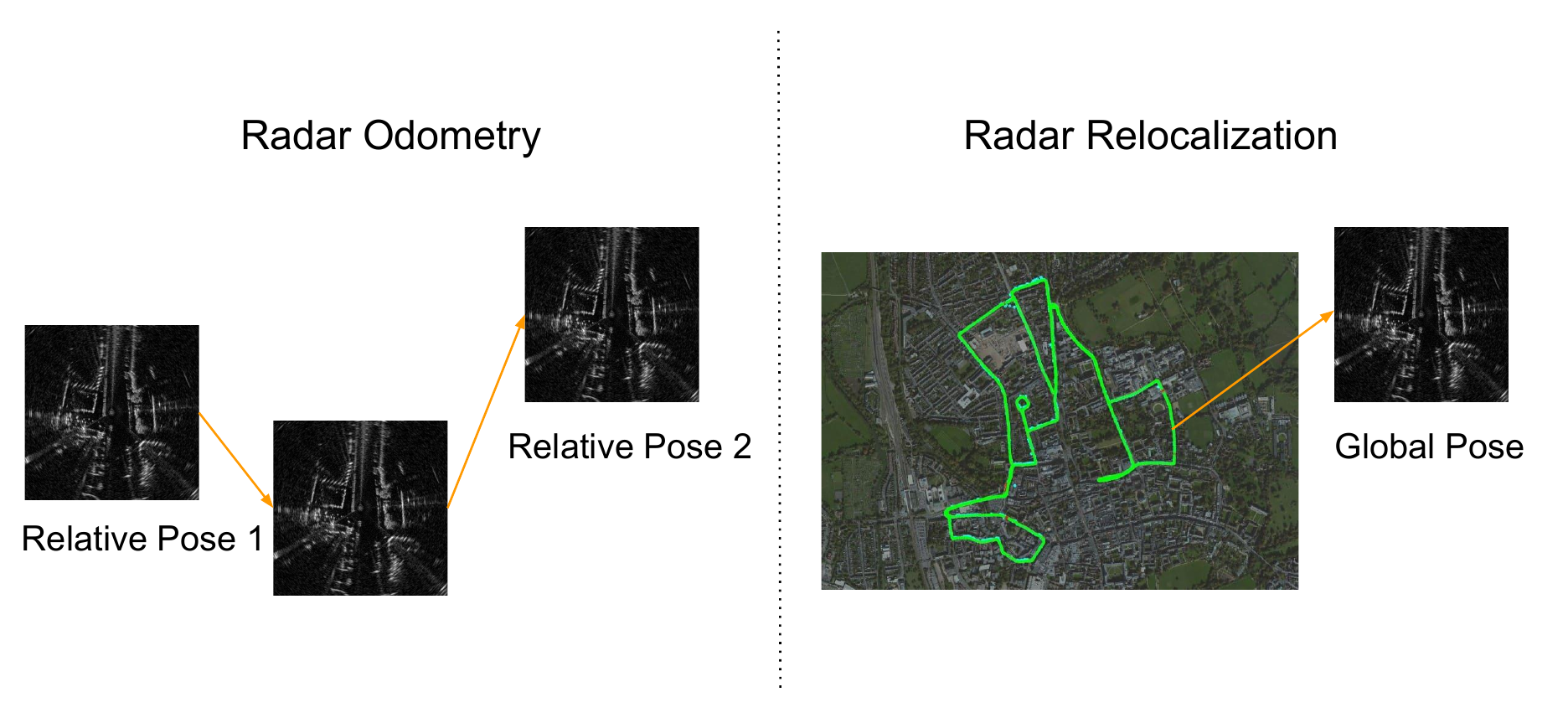}
  \caption{The difference between radar odometry and radar relocalization~\cite{Maddern2016,Barnes2020TheDataset}. Radar odometry predicts relative poses between consecutive radar scans and thus has accumulative drifts over time, while radar relocalization estimates global poses w.r.t the world coordinate and needs to traverse the environments before. These are two different tasks in localization, and this work focuses on the radar relocalization.}
  \label{fig:loc_diff}
\end{figure}

Recent works proposed to utilize radar scans for ego-motion estimation, which is known as radar odometry. Cen \textit{et al.} \cite{Cen2018PreciseConditions} extracted landmarks from radar scans and then conducted scan matching to predict ego-motion based on unary descriptors and pairwise compatibility scores. Barnes \textit{et al.}~\cite{Barnes2019MaskingInformation} developed a robust and real-time radar odometry system based on deep correlative scan matching with learnt feature embedding and self-supervised distraction-free module. Afterwards, they proposed a deep key point detection approach for radar odometry estimation and metric localization by embedding a differentiable point-based motion estimator~\cite{Barnes2020UnderRadar}. Note that different from these methods, our work focuses on radar-based absolute localization, which predicts global poses w.r.t. the world coordinate rather than relative poses. Fig.~\ref{fig:loc_diff} illustrates the differences between these two different localization tasks.

\section{Deep Radar Relocalization} \label{deep_radar_relocalization}
\begin{figure*}
  \centering
  \includegraphics[width=\linewidth]{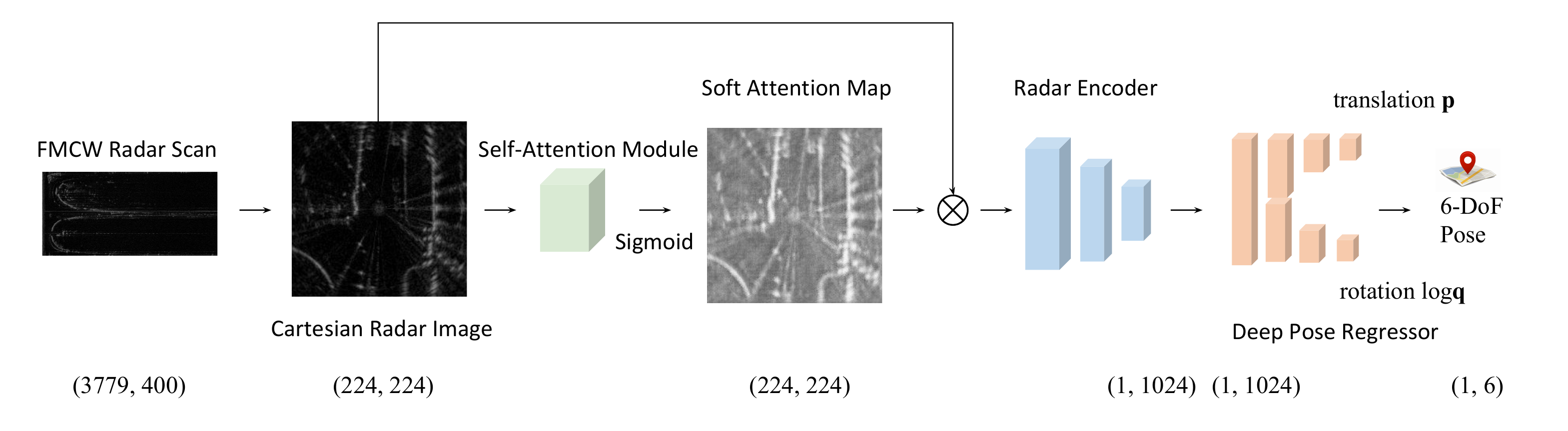}
  \caption{The architecture of RadarLoc. RadarLoc consists of a self-attention module, a radar encoder and a deep pose regressor. A raw FMCW radar scan is transformed into a Cartesian radar image, and then it is fed into a self-attention module to learn a soft attention map. DenseNet~\cite{Huang2017DenselyNetworks} is employed as the radar encoder to extract useful features for relocalization. The deep pose regressor predicts the parameterized translation $\mathbf{p} \in R^3$ and rotation $log\mathbf{q} \in R^3$ \cite{Brahmbhatt2018}. The predicted parameterized rotation vector $log\mathbf{q}$ can be further transformed to the 4-D rotation vector $\mathbf{q} \in R^4$.}
  \label{fig:radarloc_architecture}
\end{figure*}

In this section, we introduce the proposed deep radar relocalization framework in detail. The overall architecture of RadarLoc is illustrated in Fig.~\ref{fig:radarloc_architecture}, which consists of a self-attention module, a radar encoder, and a deep pose regressor. 
Since the original output of the FMCW scanning radar is a polar image, we transform it into the Cartesian space as a grey-scale birds-view-like image for better representation and improved localization performance~\cite{Wang2019Pseudo-lidarDriving}. During training phase, the neural network is optimized by the geometry-aware loss function which employs a sequence of radar scans to learn global 6-DoF poses and relative transformations simultaneously. During test phase, the RadarLoc estimates the 6-DoF pose of a single Radar input each time.
    
\subsection{Problem Formulation}
	The scope of this work is to predict absolute 6-DoF poses of the mobile agent given radar scans as inputs. The scene has been visited by the agent before, in which the agent can relocalize itself. The relocalization of the agent is parameterized by a 6-DoF pose $\mathbf{P} = [\mathbf{p}, \mathbf{q}]$ with respect to the world coordinate, where $\mathbf{p} \in R^3$ is a 3-D translation vector and $\mathbf{q} \in R^4$ is a 4-D rotation vector. At each timestamp t, the agent receives a Cartesian Radar image $\mathbf{I} \in R^{H \times W}$ from the FMCW scanning radar where $H$ is height and $W$ is width. The deep radar relocalization framework learns a function $f$ so that $f(\mathbf{I}) = [\mathbf{p}, \mathbf{q}]$, where $f$ is a deep neural network.
   
\subsection{Self-Attention for Robust Relocalization}
\begin{figure}
  \centering
  \includegraphics[width=\linewidth]{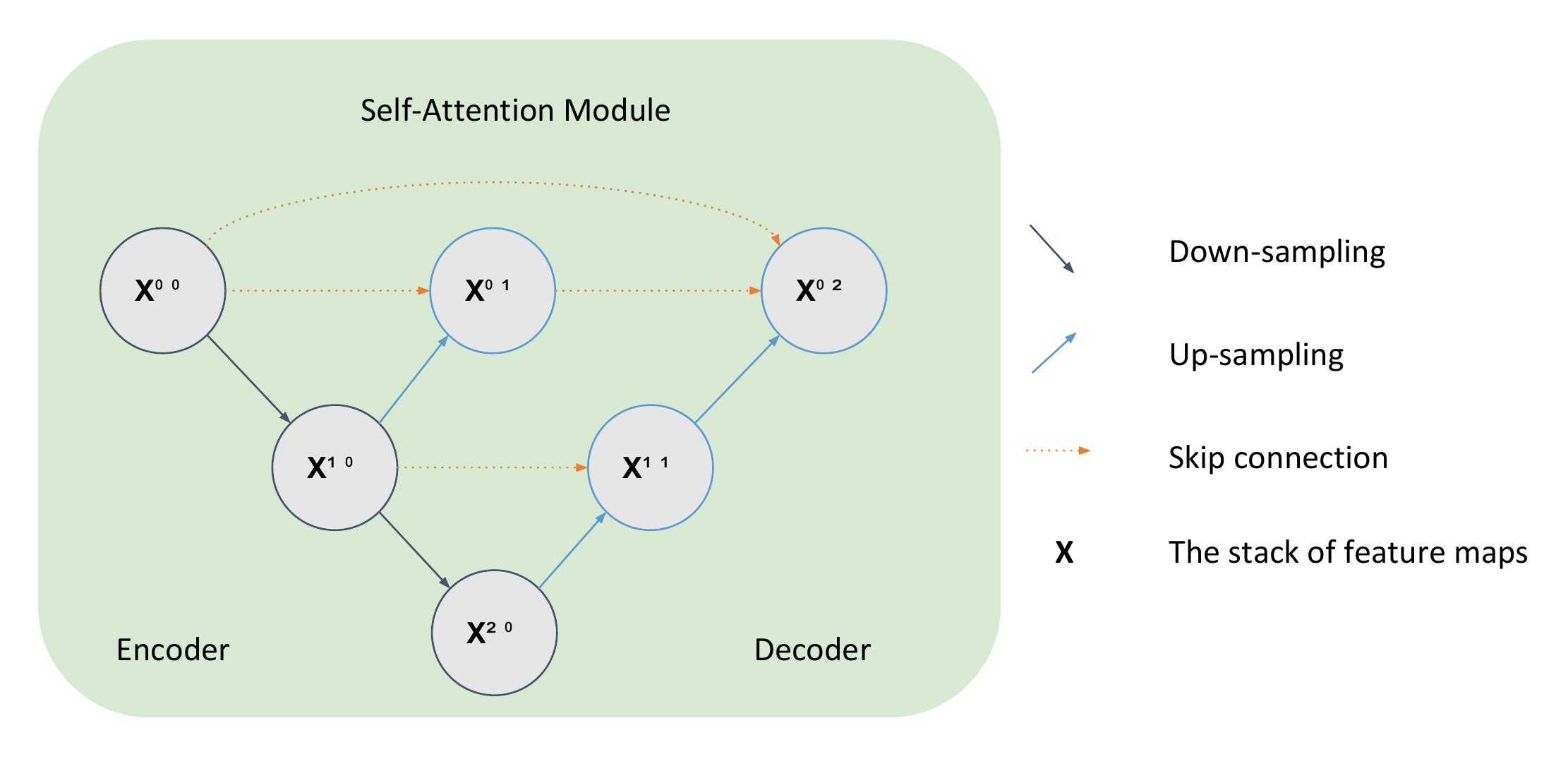}
  \caption{The architecture of self-attention module. The module is a nested encoder-decoder structure\cite{Zhou2018UNet++:Segmentation}. For better visualization, we only depict 3 levels in the figure, and in our implementation, the nested structure has 6 levels. We adopt a soft attention mechanism, and fuse output features at the upper level to have a fine-grained attention map.}
  \label{fig:radarloc_self_att}
\end{figure}

    For the radar relocalization task, there are two categories of noises which can significantly affect the accuracy of pose predictions. One is  noises from the radar sensor itself. The current FMCW scanning radar is affected by multiple noises, e.g. range error, angular error, and false positive and false negative detection which make the radar scans noisier than camera images. The other is the foreground moving objects in dynamic environments. There are several types of dynamic outliers e.g. pedestrians, bikes, buses, trucks in the complex urban environments, which have different shapes and sizes. Since the radar can scan more than 150 meters range, it is likely that one radar image can contain these different types of moving objects. Therefore, the aforementioned noises can inevitably bias the neural network, making the radar relocalization quite challenging. Barnes \textit{et al.}~\cite{Barnes2019MaskingInformation} proposed a U-Net structure to predict distraction-free radar odometry. Wang \textit{el al.}~\cite{Wang2020AtLoc:Localization} designed a non-local self-attention module to filter out moving objects for camera relocalization. However, these methods neither learn semantic features in a fine-grained manner~\cite{Barnes2019MaskingInformation} nor are designed specifically for radar images~\cite{Wang2020AtLoc:Localization}. To this end, we propose a novel self-attention module for radar relocalization as shown in Fig.~\ref{fig:radarloc_self_att}, which is a nested encoder-decoder style neural network, to mitigate the impact of these noises by filtering them out. 
Our design intuition is that considering the different shapes and sizes of moving objects, the self-attention module should have the ability to extract fine-grained features and filter out these dynamic noises. Compared to the U-Net style architecture, the nested encoder-decoder architecture can gradually down-sample, fuse and up-sample features from inputs, which can reduce the semantic gap between the feature maps and extract fine-grained semantic information. We choose the re-designed skip pathways proposed by Zhou~\textit{et al.}~\cite{Zhou2018UNet++:Segmentation} due to its impressive performance on multiple medical image segmentation tasks, in which the feature maps pass through a dense convolution block whose number of convolution layers depends on the pyramid level, and the stack of the feature maps are calculated as:
\begin{equation}
    \mathbf{X}^{i,j} = \begin{cases}
            g(\mathbf{X}^{i-1,j}),& j = 0 \\
            g([[\mathbf{X}^{i,k}]^{j-1}_{k=0}, g^{'}(\mathbf{X}^{i+1,j-1})]),& j > 0
        \end{cases}
\end{equation}
   where $\mathbf{X}^{i,j}$ is the extracted feature of each node in Fig.~\ref{fig:radarloc_self_att} , $g(*)$ denotes a convolution layer with an activation function, $g^{'}(*)$ is an up-sampling layer, $[*]$ indicates a concatenation layer, $i \in [0, 1, .., n-1]$ is the index of the down-sampling layer along the encoder, $j \in [0, 1,...,n-1]$ is the index of convolution layer of the dense block along the skip pathway, $n$ is the number of pyramid levels. In order to learn features at different scales, we fuse node outputs on the uppermost level to generate the output features $\mathbf{I}_{node}$ by averaging them:
\begin{equation}
	\mathbf{I}_{node} = \frac{1}{n}\sum_{j=0}^{n-1}\mathbf{X}^{0,j}
\end{equation}
   Thus, our self-attention module is an encoder-decoder pyramid structure with densely skip pathways followed by an activation function. We adopt a soft attention mechanism to learn the mask, so the activation function we use is \textit{Sigmoid}. Given a Cartesian radar  image $\mathbf{I} \in R^{H \times W}$, the self-attention module serves to learn a noise-free feature map $\mathbf{I}^{'} \in R^{H \times W}$:
 \begin{equation}
	\mathbf{I}^{'} = \sigma(\mathbf{I}_{node}) \cdot \mathbf{I}
\end{equation}
where $\sigma$ is the \textit{Sigmoid} function, and $\cdot$ represents the dot product.

\subsection{Radar Encoder}
	The radar encoder extracts features from a radar image for relocalization. Existing state-of-the-art camera relocalization approaches~\cite{Brahmbhatt2018,Wang2020AtLoc:Localization,Huang2019PriorEnvironments} employ ResNet~\cite{He2016} as the visual encoder considering the residual neural networks can learn deeper and alleviate the gradient vanishing problem. DenseNet~\cite{Huang2017DenselyNetworks}, which consists of densely connected convolutional networks, has been proved better performance on four object recognition tasks than ResNet. Hence, RadarLoc adopts pre-trained DenseNet as the radar encoder for feature extraction of the relocalization. We broadcast feature map $I^{'}$ to 3 channels, and replace the last 1000-dimensional fully connected layer with a M-dimensional fully connected layer. Formally, given the $\mathbf{I^{'}}$ from the self-attention module, the feature encoder $f_{encoder}$ extracts the feature vector $\mathbf{z} \in R^{M \times 1}$ from $\mathbf{I}^{'}$, which can be presented as:
\begin{equation}
	\mathbf{z} = f_{encoder}(\mathbf{I}^{'})
\end{equation}
    
\subsection{Deep Pose Regressor}
    The deep pose regressor receives the feature vector $\mathbf{z}$ from the Radar Encoder, and predicts the position $\mathbf{p}$ and the rotation $\mathbf{q}$ respectively. It consists of Multi-Layer Perceptrons (MLPs) of two branches. An activation function is applied to each layer of the MLPs except the last one. The pose regressor which ultimately estimates the global pose $\mathbf{P} = [\mathbf{p}, \mathbf{q}]$ is defined as:
\begin{equation}
	\mathbf{P} = f_{MLPs}(\mathbf{z})
\end{equation}

 \subsection{Loss Function with Geometric Constraints}
    For the loss function, we employ the definition in \cite{Brahmbhatt2018} as it has been shown to be effective in existing image-based global pose regression tasks. The vanilla loss function $h$ is defined as:
\begin{equation}
	h(\mathbf{P}, \hat{\mathbf{P}}) =\Vert \mathbf{p}-\mathbf{\hat{p}} \Vert_{1} e^{-\beta} + \beta +
    \Vert \log \mathbf{q}-\log\mathbf{\hat{q}} \Vert_{1} e^{-\gamma} + \gamma
    \label{eq:h}
\end{equation}
    where $\mathbf{p}$ and $\log\mathbf{q}$ are translation and orientation of the predicted global pose $\mathbf{P}$, $\hat{\mathbf{p}}$ and $\log\hat{\mathbf{q}}$ are translation and orientation of the ground-truth global pose $\hat{\mathbf{P}}$, $\Vert * \Vert_{1}$ denotes the $L_{1}$ loss function, $\beta$ and $\gamma$ are learnable balance factors which are initiated by $\beta^0$ and $\gamma^0$ respectively. $\mathbf{\log q}$ is the logarithmic form of a unit quaternion $\mathbf{q} = (u, \mathbf{v})$, where $u$ is a scalar and $\mathbf{v}$ is a 3-D vector, which is defined as:
\begin{equation}
    \log\mathbf{q}= \begin{cases}
            \frac{\mathbf{v}}{\Vert\mathbf{v}\Vert}\cos^{-1}u,& \text{if } \Vert\mathbf{v}\Vert \neq 0 \\
            \mathbf{0},& \text{otherwise}
        \end{cases}
\end{equation}
	
    Since a 2-D radar image can provide metric information within a wide range, we further improve the performance of relocalization by leveraging geometric constraints to optimize parameters of the neural network. 
\begin{table}
    \footnotesize
    \centering
    \caption{\small  Dataset Descriptions on the Oxford Radar RobotCar.}
    \vspace{.5em}
    \resizebox{\linewidth}{!}{
    \begin{tabular}{ccccc}
        \toprule
        \multirow{1}{*}{Scene}  & Time &  Tag & Training & Test \\
        \midrule
        Seq-01       & 2019-01-11-14-02-26 & sun & \checkmark & \\
        Seq-02       & 2019-01-14-12-05-52 & overcast & \checkmark & \\
        Seq-03       & 2019-01-14-14-48-55 & overcast & \checkmark & \\
        Seq-04       & 2019-01-15-14-24-38 & overcast & \checkmark & \\
        Seq-05       & 2019-01-18-15-20-12 & overcast & \checkmark & \\
        Seq-06       & 2019-01-10-11-46-21 & rain & & \checkmark \\
        Seq-07       & 2019-01-14-12-41-28 & overcast & & \checkmark \\
        Seq-08       & 2019-01-15-13-06-37 & overcast & & \checkmark \\
        Seq-09       & 2019-01-17-14-03-00 & sun & & \checkmark \\
        Seq-10       & 2019-01-18-14-14-42 & overcast & & \checkmark \\
        \bottomrule
    \end{tabular}
    }
    \label{tab:dataset_radar_robotcar}
    \vspace{-.5em}
\end{table}
During training, we choose N radar images, consisting of the current radar image $I_{0}$ as well as N-1 sequential radar images $\{I_{1},...,I_{N-1}\}$ close to $I_{0}$. Consequently, RadarLoc learns both global poses ($\mathcal{L}_{gp}$) and relative pose transformations ($\mathcal{L}_{rp}$) between radar image pairs. The improved loss functions are defined as:    
\begin{equation}
	\mathcal{L}_{gp} = \sum_{i=0}^{N-1} h(\mathbf{P}_{i}, \hat{\mathbf{P}_{i}})
	\quad \mathcal{L}_{rp} = \sum_{i=0}^{N-2}h(\mathbf{Q}_{i}, \hat{\mathbf{Q}_{i}})
\end{equation}
    where $\mathbf{P}_{i},\mathbf{Q}_{i}$ are predicted global poses and relative pose transformations while $\hat{\mathbf{P}_{i}},\hat{\mathbf{Q}_{i}}$ are ground-truth global poses and relative pose transformations respectively, and $h$ is the distance function defined in Eq.~\ref{eq:h}. Therefore, the ultimate loss function for RadarLoc is formulated as:
\begin{equation}
	\mathcal{L}_{total} = \mathcal{L}_{gp} + \mathcal{L}_{rp}
\end{equation}
Importantly, we employ multiple images in the training phase, and only a single radar image in the test phase.
\section{Experiments} \label{experiments}

In this section, we evaluate our proposed RadarLoc on the recently released Oxford Radar RobotCar Dataset~\cite{Barnes2020TheDataset,Maddern2016}, and compare it with state-of-the-art radar-based localization and deep camera and LiDAR relocalization methods.
\subsection{Dataset}
The Dataset provides Navtech CTS350-X FMCW scanning radar data, RGB images and corresponding ground truth poses. It was collected in January 2019 over thirty-two traversals of a central Oxford route spanning a total of 280 km of urban driving, and covered different kinds of lighting, weather and traffic conditions~\cite{Barnes2020TheDataset}. The length of each sequence is around 9 km, and they traverse the same route. Therefore, the dataset is large-scale and complex. For the relocalization task, it is quite challenging since the urban scenes encompass a variety of foreground objects e.g. people, car, bus, which significantly influence the performance of relocalization. The descriptions of our training sequences and test sequences from the Oxford Radar RobotCar Dataset are illustrated in Table~\ref{tab:dataset_radar_robotcar}. Note that seasonal variations affect localization significantly, this dataset only covers January.

\begin{table*}
    \footnotesize
    \centering
    \caption{\small Results showing the mean translation error (m) and rotation error (\degree) for state-of-the-art radar-based localization methods and deep camera and LiDAR relocalization methods on the Oxford Radar RobotCar Dataset. For RadarSLAM and Adapted methods, the sensory data is FMCW radar scan. The sensory data of AtLoc and PointLoc are camera RGB image and LiDAR point cloud respectively.}
    \vspace{.5em}
    \resizebox{\linewidth}{!}{
    \begin{tabular}{cccccccc|c}
        \toprule
        \multirow{3}{*}{Sequence}   & RadarSLAM  & Adapted Masking & Adapted PoseNet17  & Adapted AtLoc & Adapted LSTM & AtLoc & PointLoc & RadarLoc \\
                & \cite{Hong2020RadarSLAM:Weathers} & \cite{Barnes2019MaskingInformation} & \cite{Kendall2017} & \cite{Wang2020AtLoc:Localization} & \cite{Walch2017} & \cite{Wang2020AtLoc:Localization} & \cite{Wang2020} & (Ours) \\
                & [Radar] & [Radar] & [Radar] & [Radar] & [Radar] & [RGB] & [LiDAR] & [Radar] \\
        \midrule
        Seq-06       &49.81m, 5.22\degree & 12.54m, 3.93\degree & 15.12m, 4.08\degree & 15.85m, 4.20\degree & 15.86m, 4.28\degree & 15.36m, 3.37\degree & 14.42m, {\bf 2.77\degree} & {\bf 8.43m}, 3.44\degree \\
        Seq-07       &24.73m, 3.36\degree & 8.11m, 3.04\degree & 13.59m, 3.54\degree & 13.23m, 3.82\degree & 13.33m, 2.47\degree & 39.76m, 8.31\degree & 8.46m, {\bf 1.82 \degree} & {\bf 5.12m}, 2.87\degree \\
        Seq-08       &26.09m, 1.57\degree & 11.32m, 4.18\degree & 14.81m, 3.46\degree & 14.17m, 2.94\degree & 14.86m, 2.88\degree & 31.68m, 4.34\degree & 9.52m, {\bf 2.14\degree} & {\bf 6.56m}, 3.06\degree \\
        Seq-09       & 39.84m, 5.67\degree & 11.53m, 2.76\degree & 14.44m, 3.04\degree & 15.71m, 3.23\degree & 13.86m, 2.71\degree & 47.06m, 9.38\degree & 11.52m, {\bf 1.98\degree} & {\bf 6.51m}, 2.91\degree \\
         Seq-10       & 17.83m, 1.71\degree & 9.42m, 1.81\degree & 13.21m, 2.02\degree & 13.22m, 1.94\degree & 14.65m, 1.89\degree & 10.35m, {\bf 1.26\degree} & 8.43m, 1.40\degree & {\bf 5.34m}, 1.78\degree \\
        \midrule
        Average    & 31.66m, 3.50\degree & 10.58m, 3.15\degree & 14.23m, 3.23\degree & 14.44m, 3.22\degree & 14.51m, 2.85\degree & 28.84m, 5.33\degree & 10.47m, {\bf 2.02\degree} & {\bf 6.39m}, 2.81\degree  \\
        \bottomrule
    \end{tabular}
    }
    \label{tab:results}
    \vspace{-.5em}
\end{table*}

\subsection{Implementation}
The spatial dimensions of the self-attention module of RadarLoc are 8, 16, 32, 64, 128 and 256 respectively. The size of a Cartesian radar image is set to $224 \times 224$ in order to utilize the pre-trained DenseNet on ImageNet. For all experiments, the number of training epochs is set to 100, and we tune all baseline methods for the best performance. The learning rate is set to $1 \times 10^{-4}$, and we set the initial values of $\beta_{0} = 0.0$ and $\gamma_{0} = -3.0$.  Furthermore, we retrieve a sequence of $N=4$ radar images each time. For all methods, Adam~\cite{Kingma2015} optimizer is applied to the neural networks. 

\subsection{Baselines}

We compare RadarLoc with both radar-based methods and state-of-the-art RGB and Lidar techniques. We also adapt learning-based visual relocalization pipelines to use radar images as input. RadarSLAM is a full radar-based graph SLAM system for reliable localization in large-scale scenarios. Masking by Moving~\cite{Barnes2019MaskingInformation} is the state-of-the-art deep learning-based radar odometry approach, and we adapt the feature extraction module for relocalization. PoseNet17~\cite{Kendall2017}, LSTM-Pose~\cite{Walch2017}, and AtLoc~\cite{Wang2020AtLoc:Localization} are state-of-the-art camera image-based deep relocalization methods, and since our radar scan can be seen as a 2-D $224 \times 224$ grey-scale image, we want to examine the performance of these architectures on the radar inputs. We apply these neural networks to radar images for adapted radar relocalization. AtLoc (RGB) is the state-of-the-art deep camera relocalization method, and PointLoc (LiDAR) is the state-of-the-art DNN-based LiDAR point cloud relocalization method.

\subsection{Results}
\begin{figure*}
    \begin{subfigure}
        \centering
        \includegraphics[width=.24\textwidth]{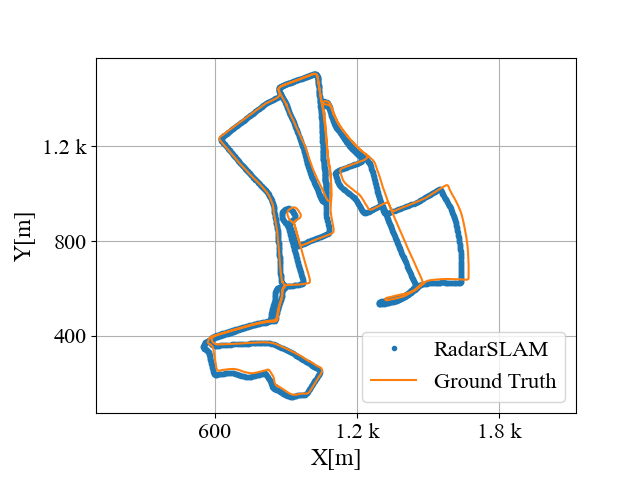}
    \end{subfigure}
    \begin{subfigure}
        \centering
        \includegraphics[width=.24\textwidth]{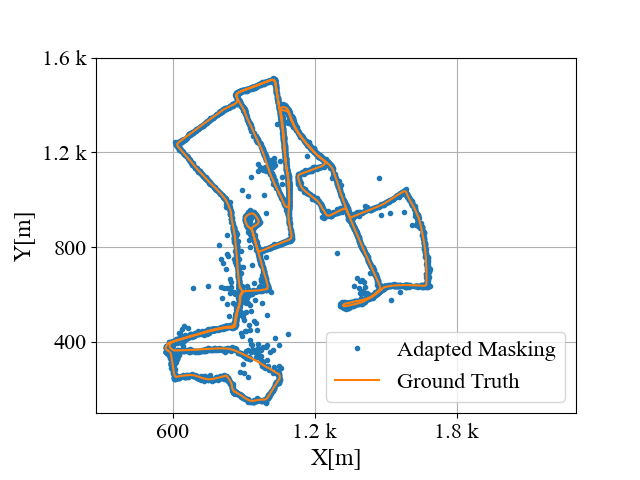}
    \end{subfigure}
    \begin{subfigure}
        \centering
        \includegraphics[width=.24\textwidth]{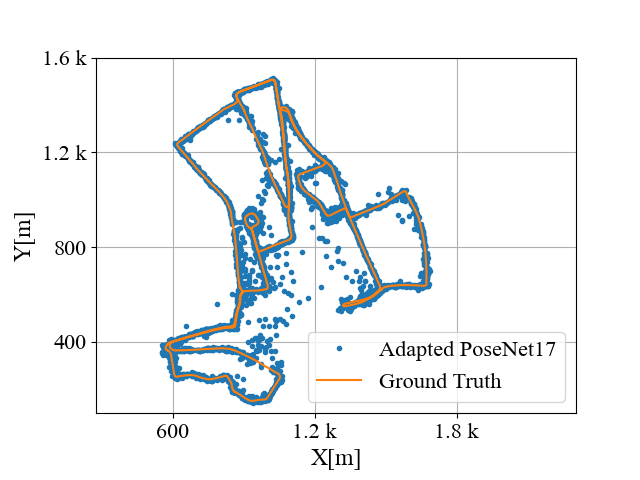}
    \end{subfigure}
     \begin{subfigure}
        \centering
        \includegraphics[width=.24\textwidth]{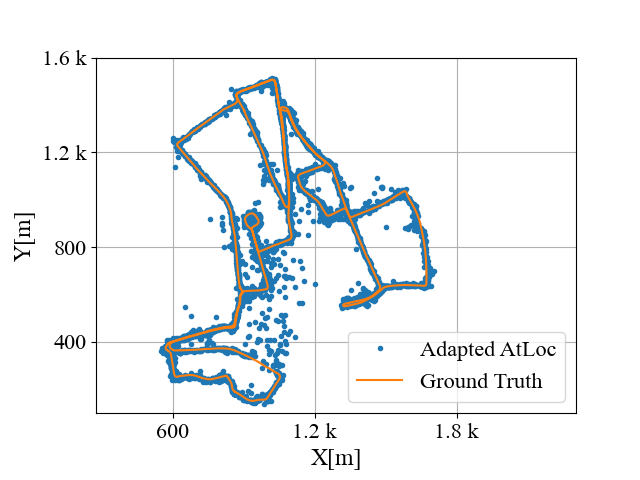}
    \end{subfigure}
    \newline
     \begin{subfigure}
        \centering
        \includegraphics[width=.24\textwidth]{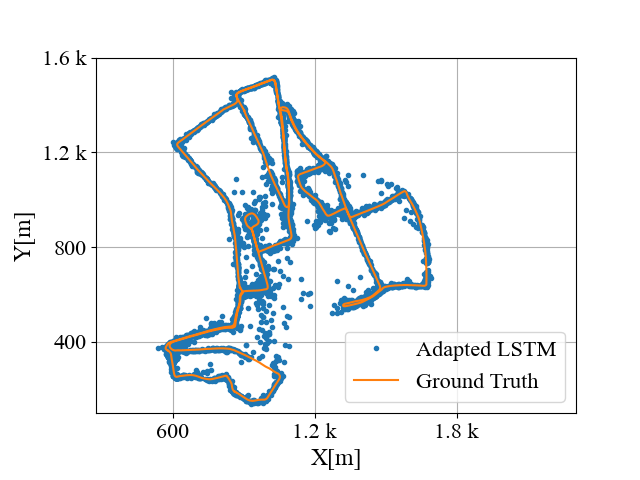}
    \end{subfigure}
    \begin{subfigure}
        \centering
        \includegraphics[width=.24\textwidth]{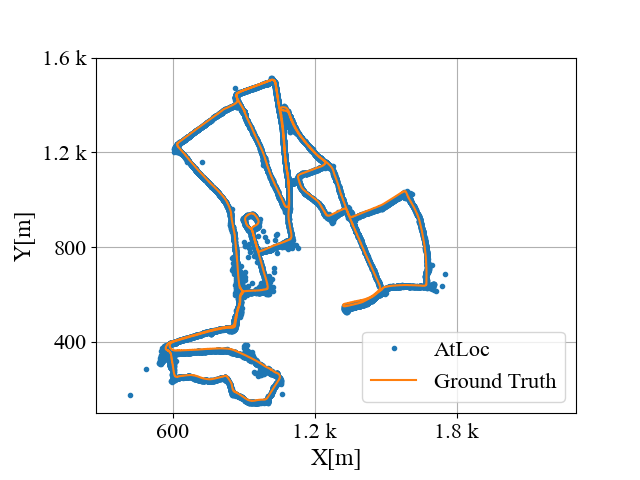}
    \end{subfigure}
    \begin{subfigure}
        \centering
        \includegraphics[width=.24\textwidth]{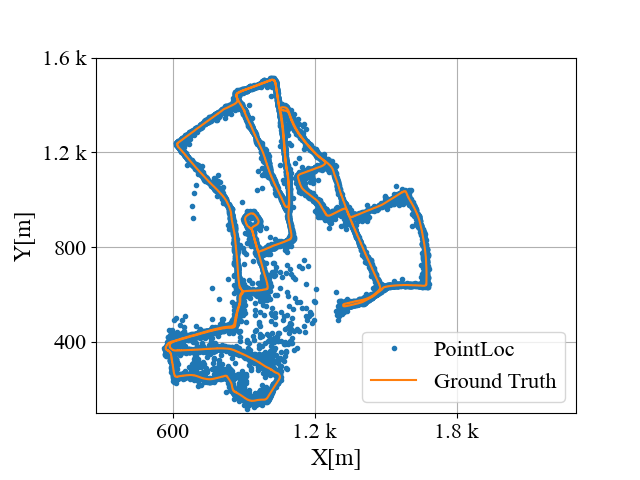}
    \end{subfigure}
    \begin{subfigure}
        \centering
        \includegraphics[width=.24\textwidth]{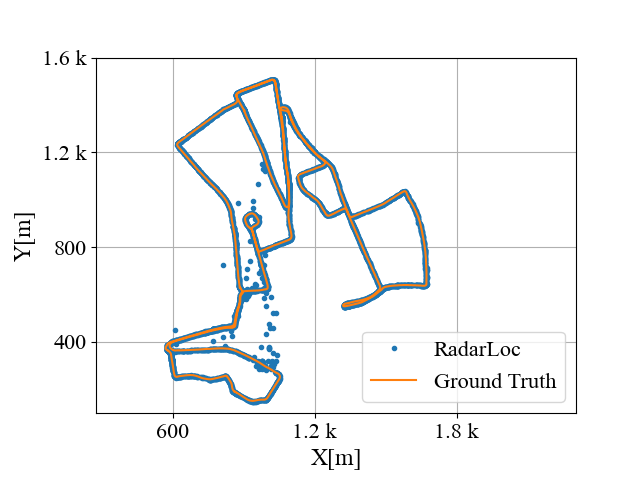}
    \end{subfigure}
    \caption{Visual comparisons of all localization approaches for Sequence 10. Poses were projected from 6-DoF to 3-DoF with exception to RadarSLAM, which outputs 3-DoF poses originally. }
    \label{fig:paths_gt}
\end{figure*}

The experimental results are illustrated in Table~\ref{tab:results}, and the qualitative comparisons are depicted in Fig.~\ref{fig:paths_gt}. From Table~\ref{tab:results}, the proposed RadarLoc outperforms radar-based methods by a significant margin. RadarSLAM can predict consecutive poses but accumulates drifts with the increasing distance, which leads to large localization errors as shown in Table~\ref{tab:results}. Note also that RadarSLAM is a continuous localization technique, while RadarLoc is single-shot. For adapted Masking by Moving, adapted PoseNet17, adapted AtLoc and adapted LSTM, the results indicate that our proposed neural network architecture is superior than previously proposed architectures for both deep radar odometry and deep camera relocalization. 

\begin{figure*}
    \centering
    \begin{tabular}{c c c c c}
        \includegraphics[width=.18\textwidth]{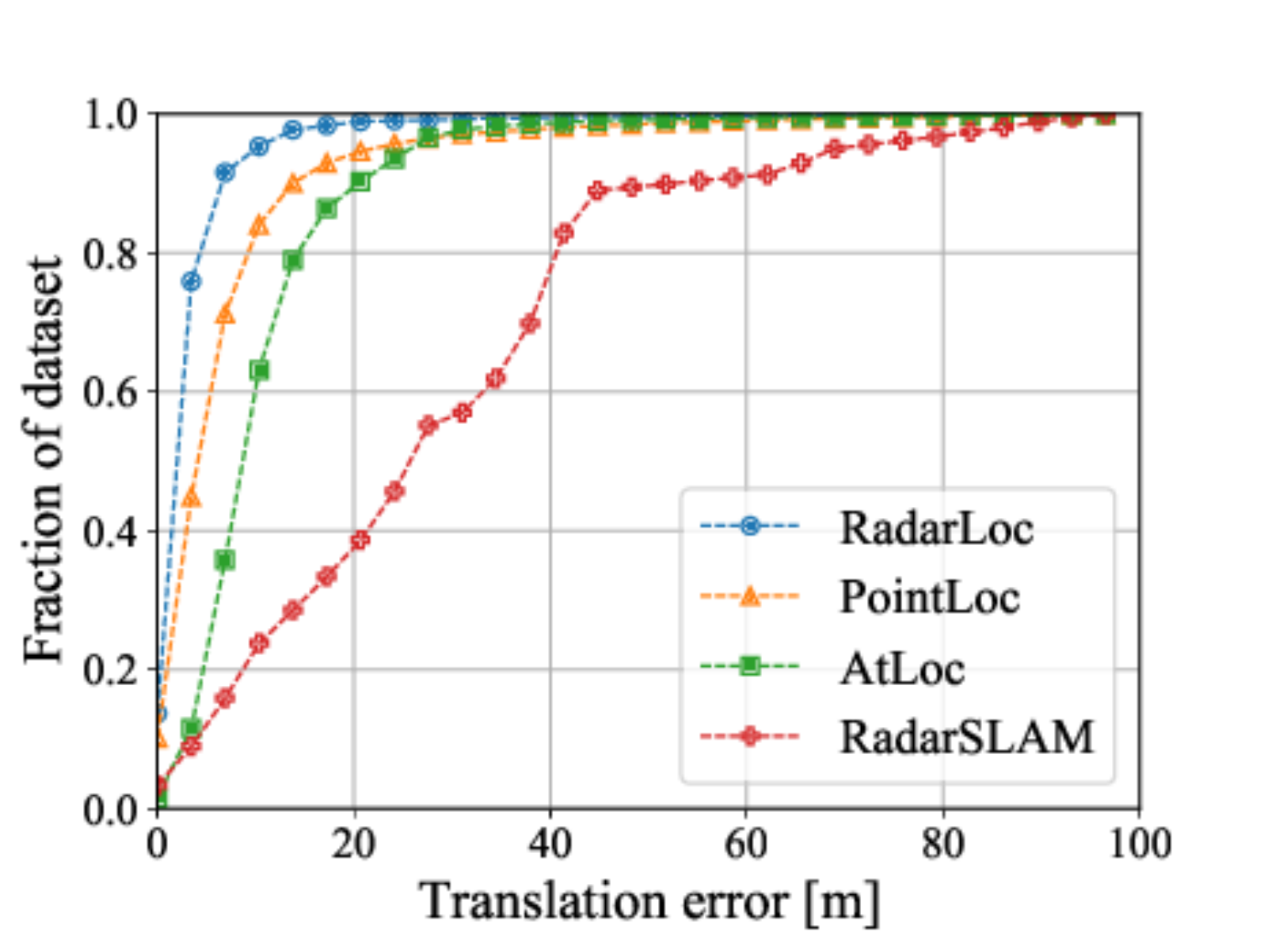} &
        \includegraphics[width=.18\textwidth]{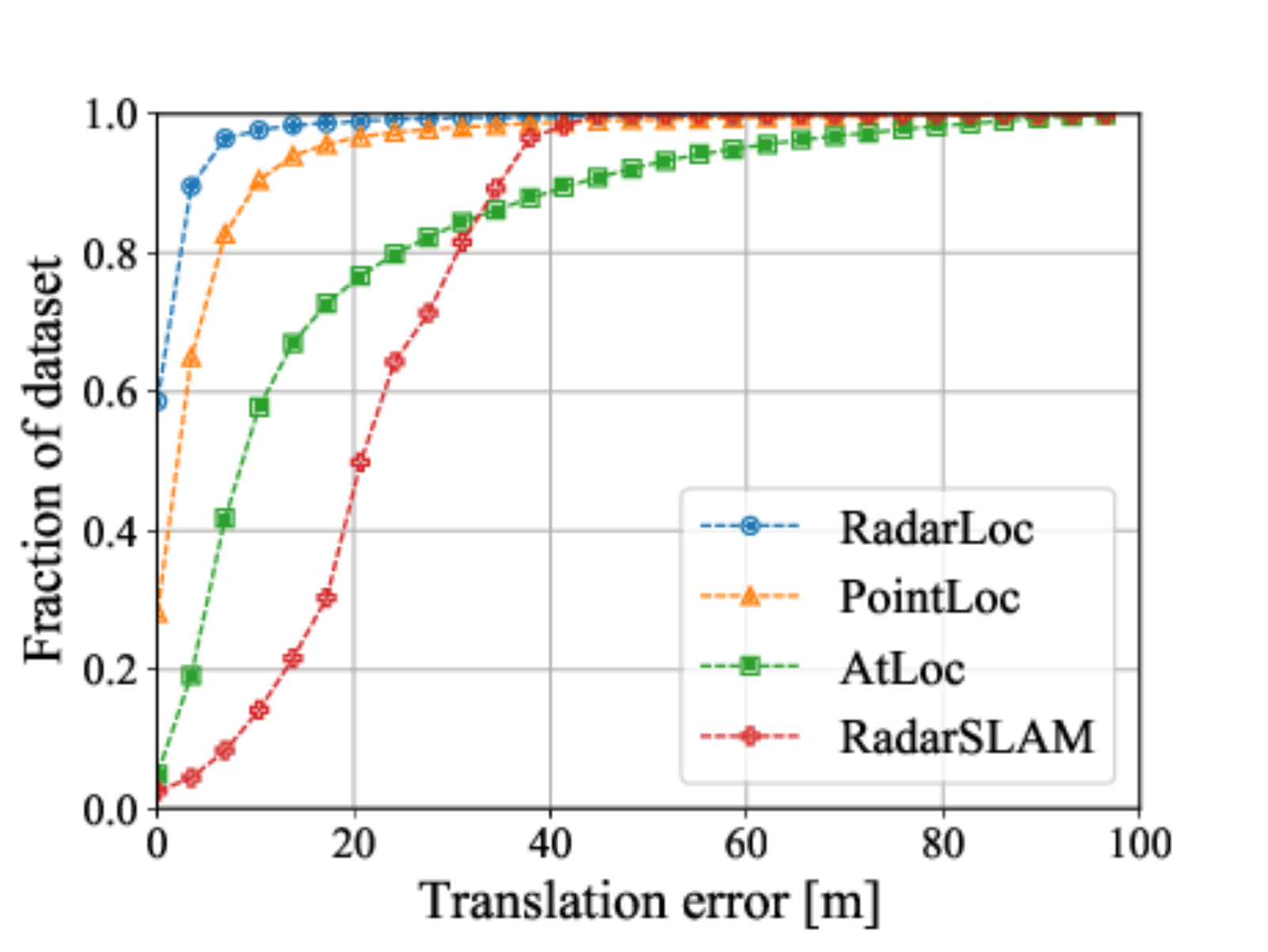} &
        \includegraphics[width=.18\textwidth]{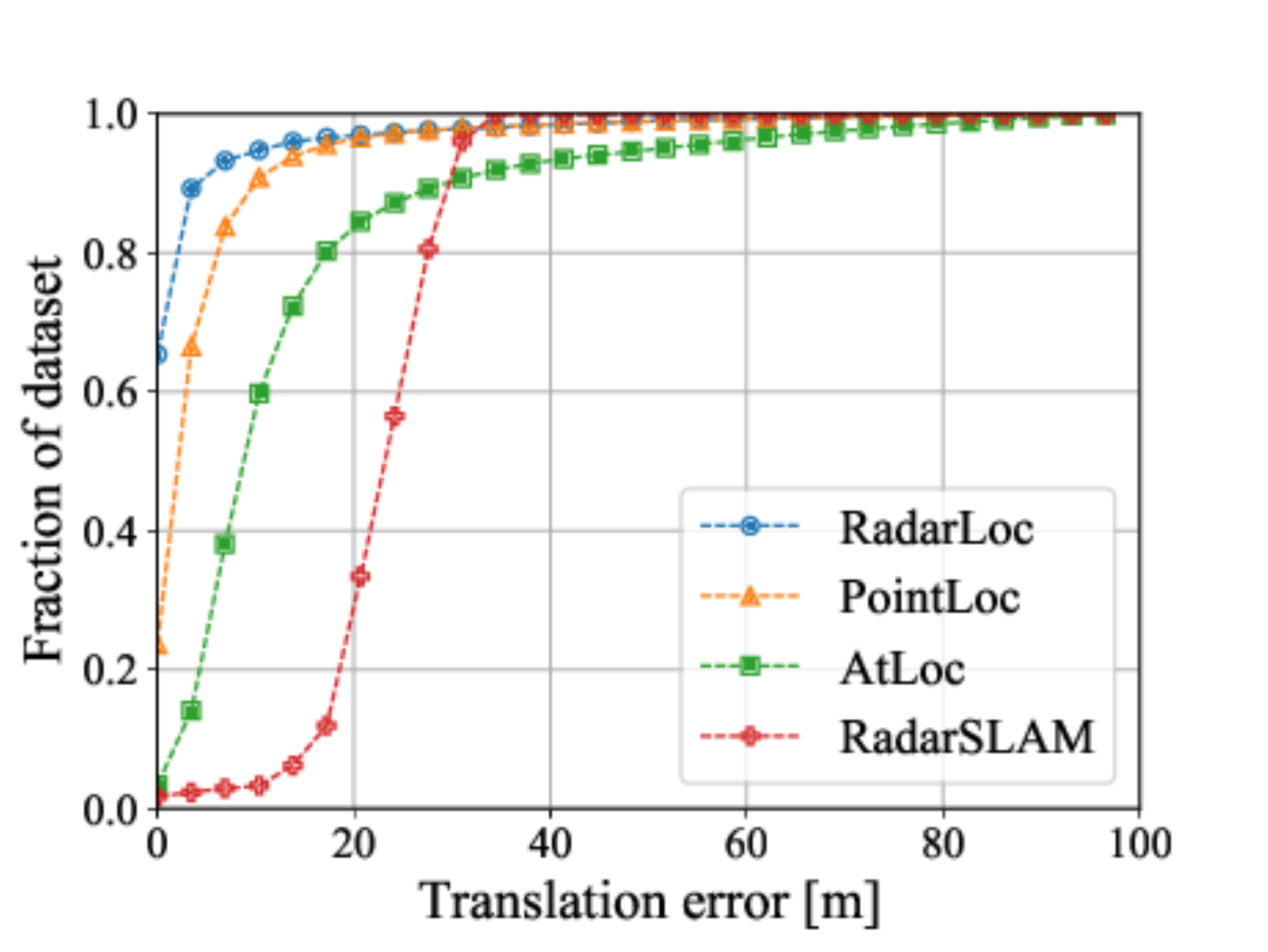} &
        \includegraphics[width=.18\textwidth]{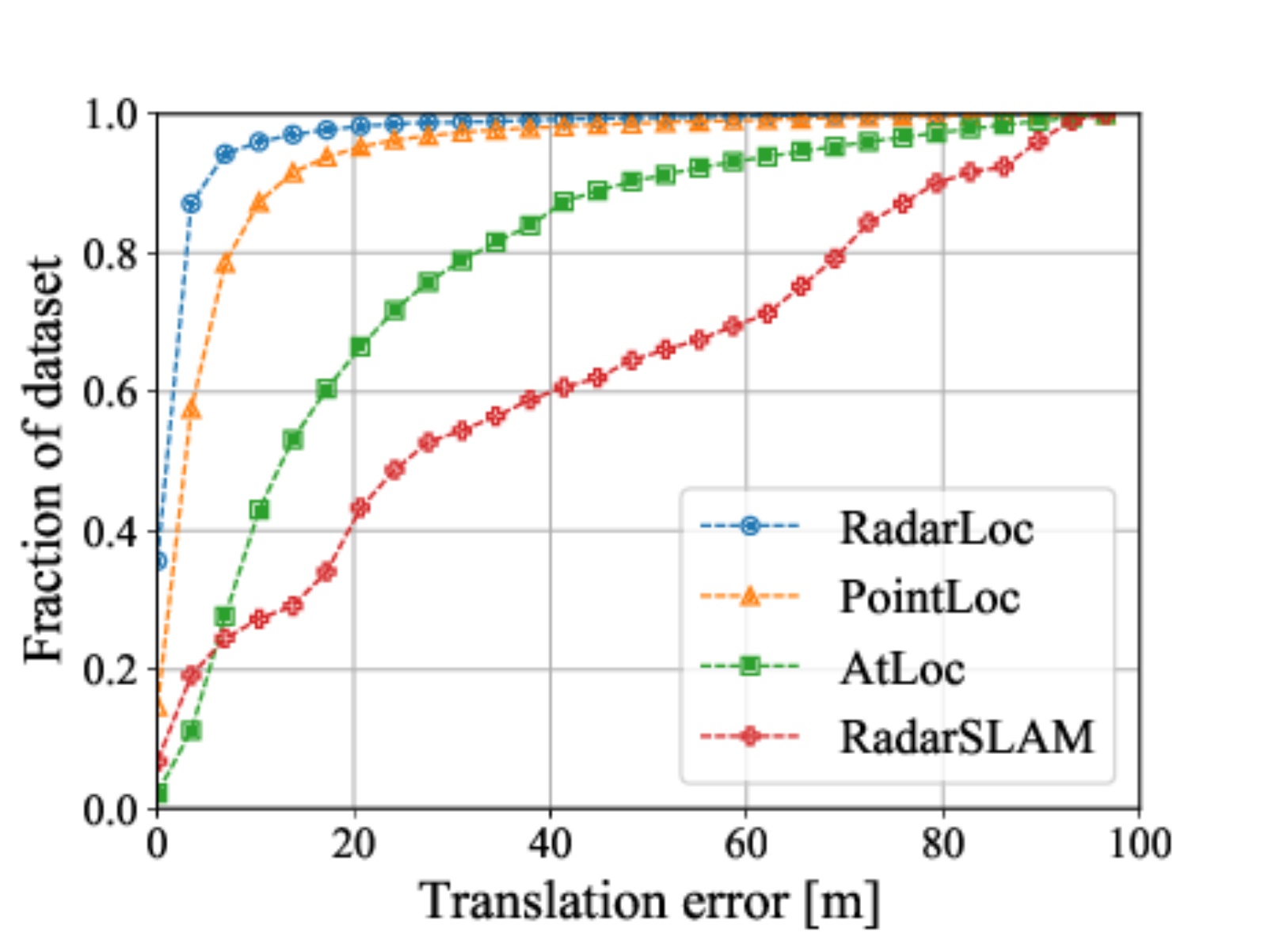} &
        \includegraphics[width=.18\textwidth]{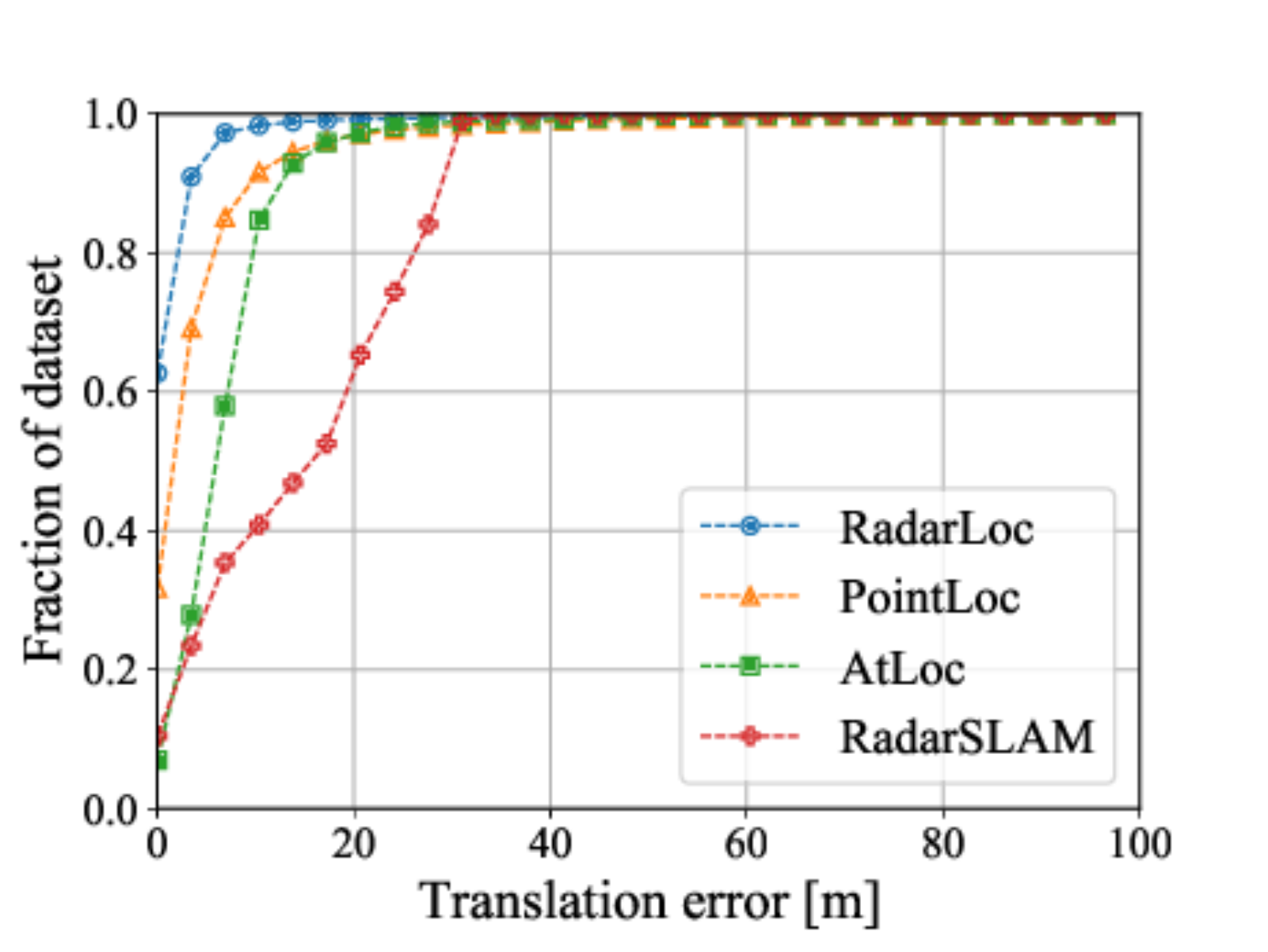} \\
        \includegraphics[width=.18\textwidth]{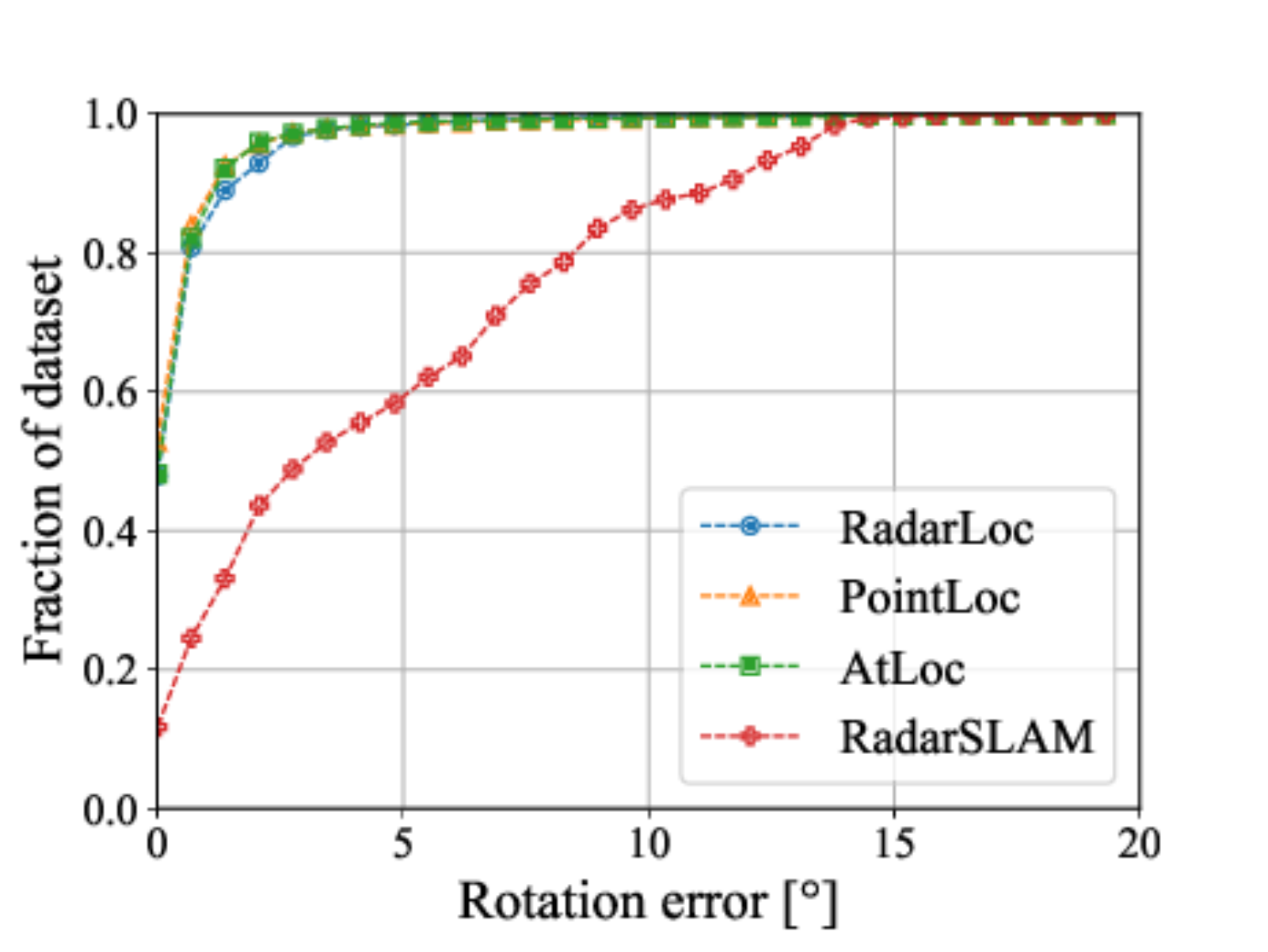} &
        \includegraphics[width=.18\textwidth]{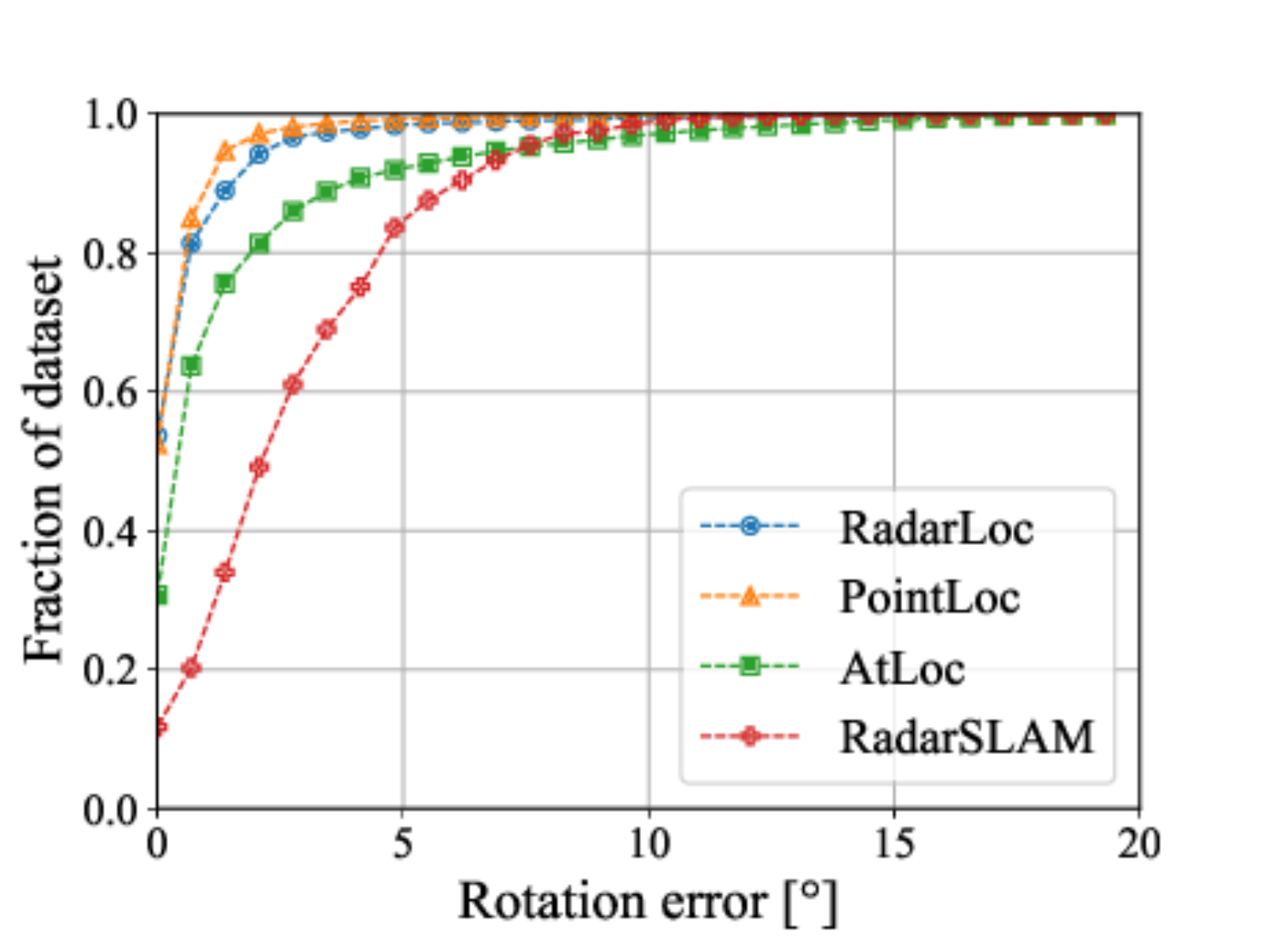} &
        \includegraphics[width=.18\textwidth]{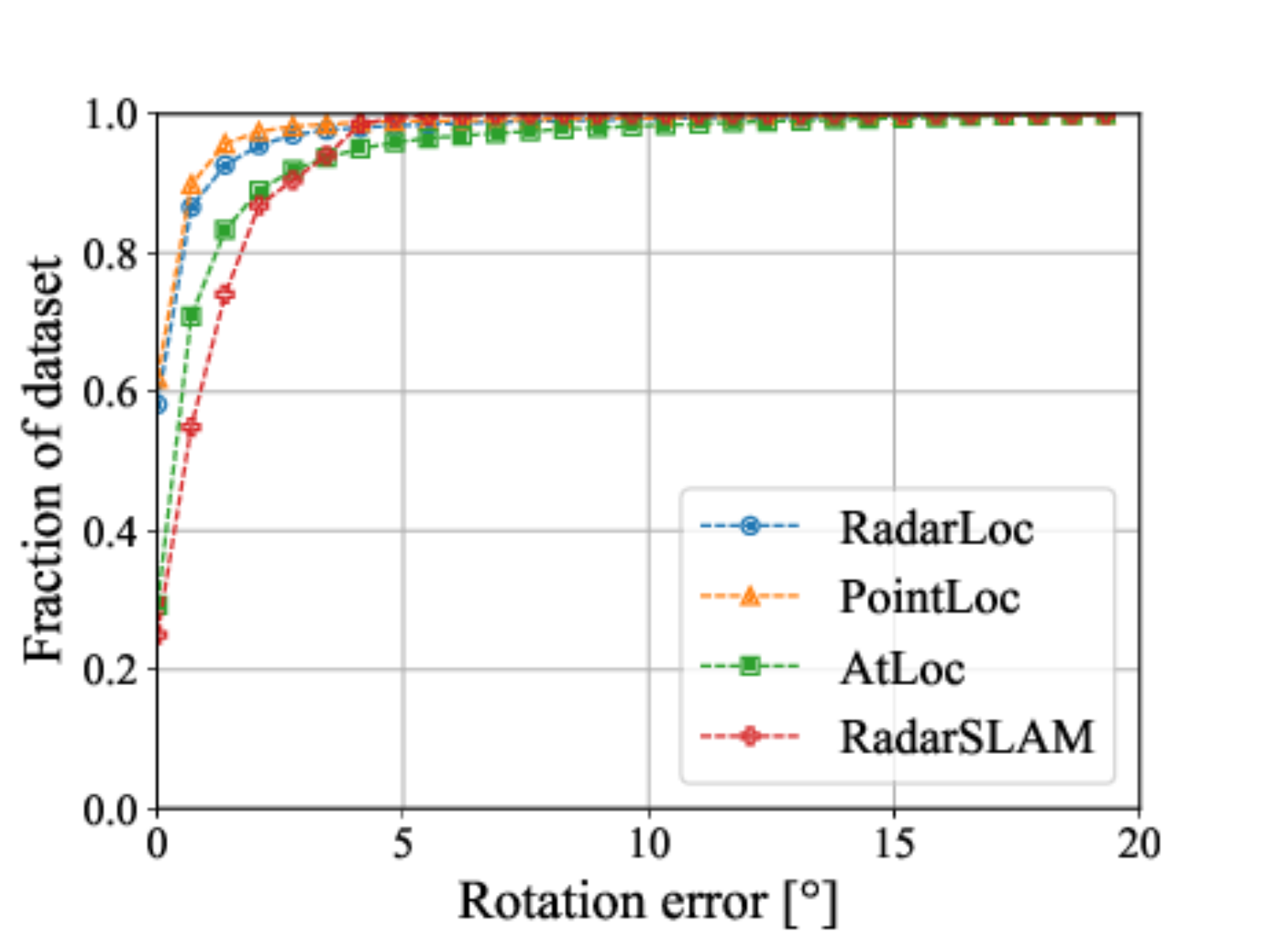} &
        \includegraphics[width=.18\textwidth]{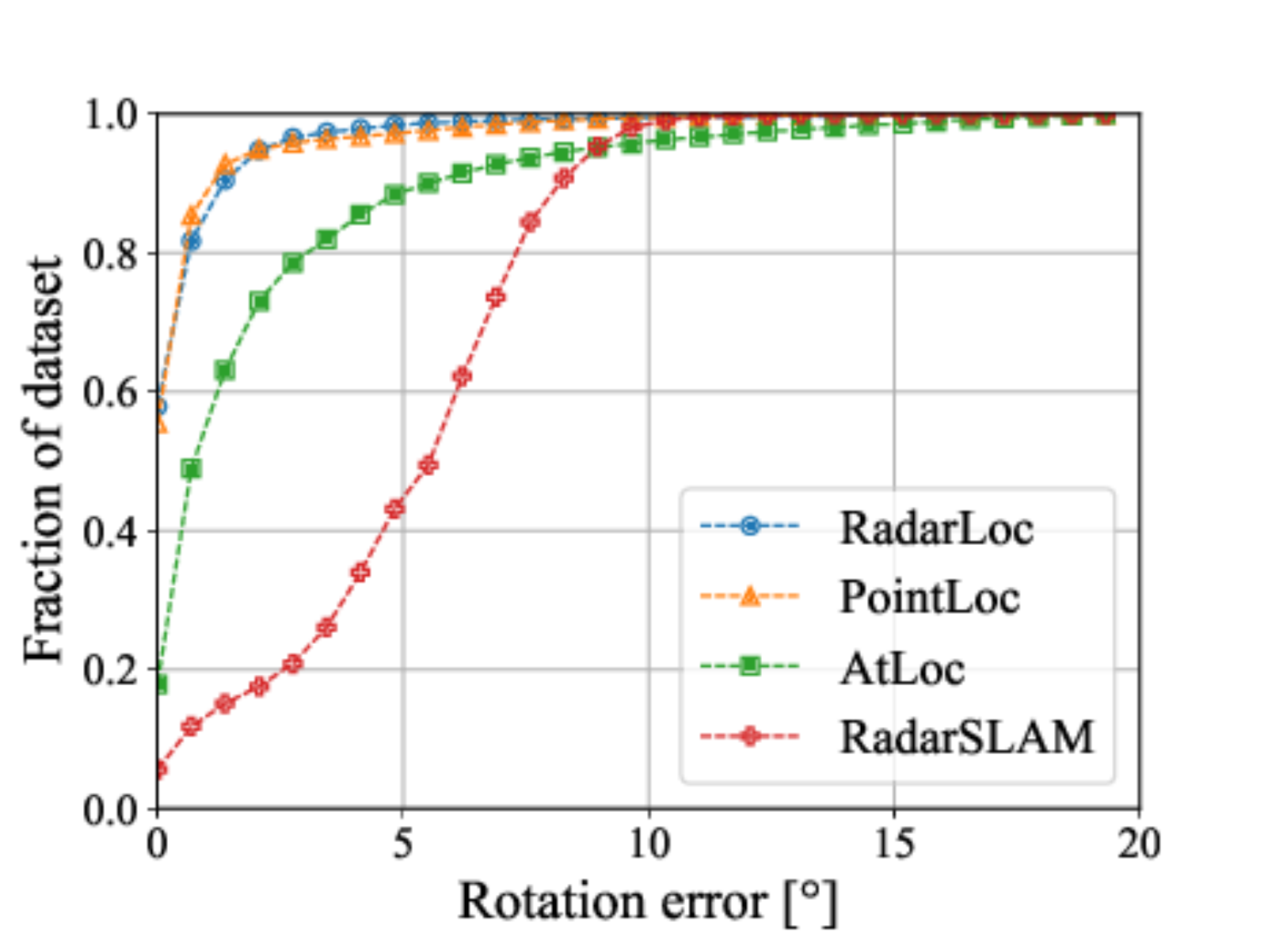} &
        \includegraphics[width=.18\textwidth]{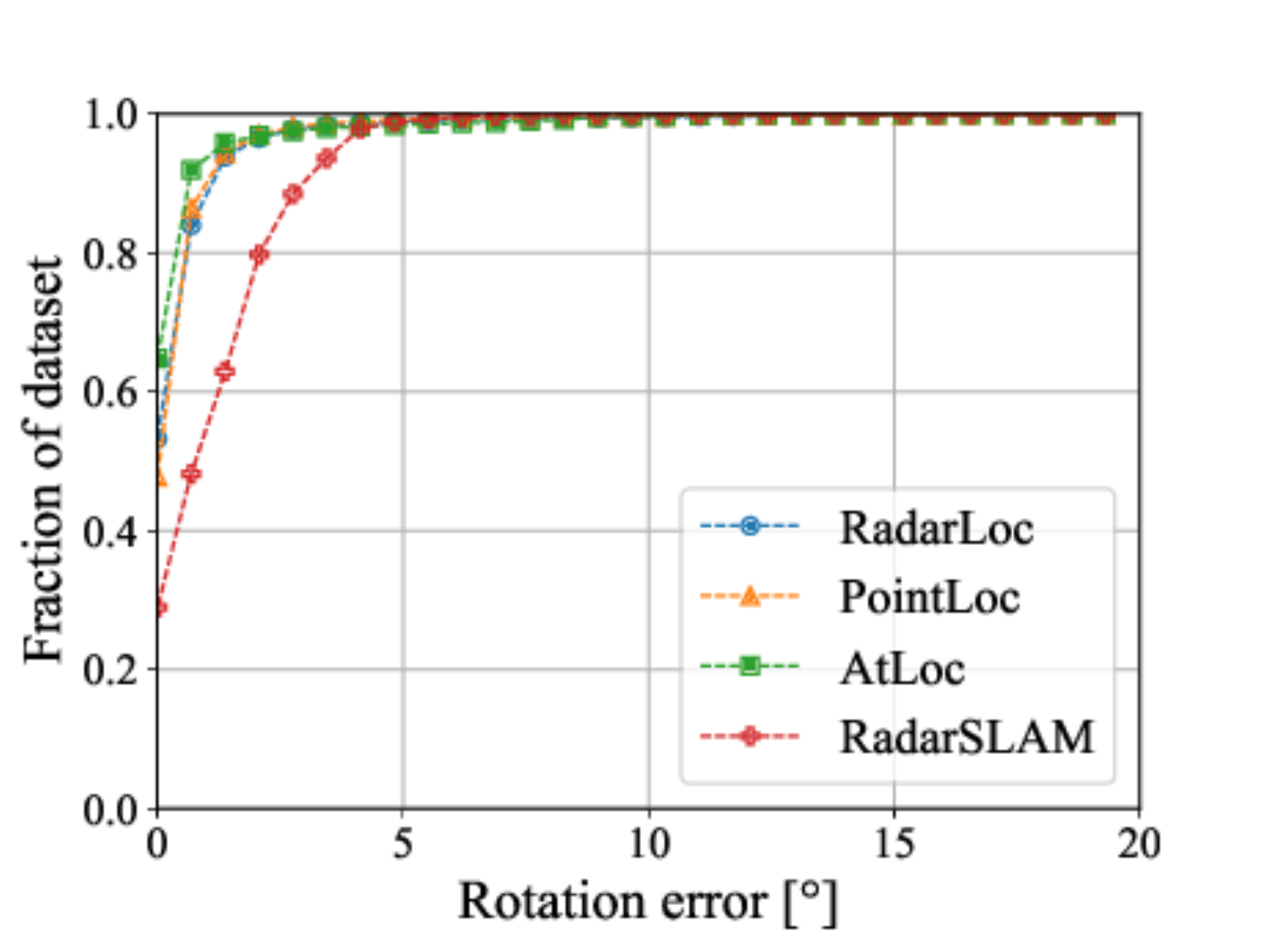} \\
        \small (a) Sequence 06 & 
        \small (b) Sequence 07 &
        \small (c) Sequence 08 &
        \small (d) Sequence 09 &
        \small (e) Sequence 10 
    \end{tabular}
    \medskip
    \caption{Cumulative distributions of translation and rotation errors.}
    \label{fig:cdf_trans_rot}
\end{figure*}

We also compare with camera-based and LiDAR-based deep relocalization methods to examine the differences among different sensory data for deep pose regression.  The results in Table~\ref{tab:results} demonstrate that radar-based deep relocalization method is much better than camera-based method in terms of accuracy. The probable reasons are radar sensors can provide broader FoV of scenes and are less sensitive to environmental conditions than cameras. Interestingly, RadarLoc significantly outperforms PointLoc in translation while remaining comparable in rotation performance. This is most likely due to LiDAR providing full 3-D metric depth, rather than the 2-D bird's eye view of the FMCW radar scanner, which aids in full 6-DoF pose estimation.

Fig.~\ref{fig:cdf_trans_rot} shows the cumulative distribution function (CDF) for both translation and orientation errors for the above mentioned approaches. RadarLoc consistently produces low errors in all sequences and it is closely followed by PointLoc. Pose accuracy for RadarSLAM and AtLoc, on the other hand, was highly dependent on the sequence being considered. Most noticeably, in Sequence 09, over 40\% of poses estimated with RadarSLAM showed errors beyond 40m.   

\subsection{Ablation Study}
In order to study the impact of different components of the proposed RadarLoc system, we conduct the ablation study as shown in Table~\ref{tab:ablation}. For ablation experiments, we keep all the architecture designs the same as RadarLoc except that we do not contain self-attention module (w/o SA), use the UNet as the self-attention module (SA w/ UNet), use the ResNet as the radar encoder (ResNet) and do not use the geometric constraints as one component of the loss function (w/o GC) respectively. The RadarLoc improves the w/o SA by 39.77\% in translation and 5.70\% in rotation, which proves that our self-attention module is very effective in improving the radar localization performance. To delve into the reasons behind the improvement of our self-attention module, we visualize the soft attention map as depicted in Fig.~\ref{fig:radarloc_architecture}. The self-attention module helps RadarLoc focus more on the static objects like streets and buildings rather than feature-less regions of a radar image. Moreover, it improves the SA w/ UNet by 12.71\%  and the ResNet by 31.51\% in translation while remains comparable performance in rotation (less than $0.1\degree$). Note that the performance of translation is very crucial in application scenarios like indoor parking lot or outdoor autonomous robots. Furthermore, the RadarLoc improves the w/o GC by 36.63\% in translation and 5.39\% in rotation, which demonstrates that the geometric constraints can greatly improve the performance of radar relocalization. Meanwhile, RadarLoc without geometric constraints (w/o GC) also outperforms all the baselines in Table~\ref{tab:results} except the rotation of PointLoc, which reveals the effectiveness of the proposed neural network architecture considering the only difference between RadarLoc and the w/o GC is the loss function.

\begin{table}
    \footnotesize
    \centering
    \caption{\small Results showing the mean translation error (m) and rotation error (\degree) of ablation studies on the Oxford Radar RobotCar Dataset.}
    \vspace{.5em}
    \resizebox{\linewidth}{!}{
    \begin{tabular}{ccccc|c}
        \toprule
        Sequence   & w/o SA & SA w/ UNet & ResNet & w/o GC & RadarLoc \\
        \midrule
        Seq-06       & 12.56m, 3.89\degree & 9.96m, 3.62\degree & 11.51m, 3.62\degree & 11.13m, 3.80\degree & {\bf 8.43m}, {\bf3.44\degree} \\
        Seq-07       & 10.26m, 3.16\degree & 6.74m, 2.76\degree & 8.09m, {\bf2.75\degree} & 8.04m, 2.95 \degree & {\bf 5.12m}, 2.87\degree \\
        Seq-08       & 10.91m, 3.38\degree & {\bf6.46m}, 2.72\degree & 9.42m, {\bf2.60\degree} & 10.74m, 3.47\degree & 6.56m, 3.06\degree \\
        Seq-09       & 10.36m, 2.82\degree & 7.77m, 3.13\degree & 9.73m, 2.86\degree & 10.34m, {\bf 2.77\degree} & {\bf 6.51m}, 2.91\degree \\
         Seq-10      & 8.94m, 1.65\degree & 5.64m, {\bf1.61\degree} & 7.91m, 1.81\degree & 10.15m, 1.88\degree & {\bf 5.34m}, 1.78\degree \\
        \midrule
        Average    & 10.61m, 2.98\degree & 7.32m, 2.77\degree & 9.33m, {\bf2.73\degree} & 10.08m, 2.97\degree & {\bf 6.39m}, 2.81\degree  \\
        \bottomrule
    \end{tabular}
    }
    \label{tab:ablation}
    \vspace{-.5em}
\end{table}

\section{CONCLUSIONS} ~\label{conclusions}
The paper proposes a novel Radar-based relocalization system, RadarLoc, based on deep learning. It can directly predict 6-DoF global poses in an end-to-end fashion. 
The system can be leveraged in urban areas like Oxford for localization or as a component of the existing radar localization system to redeem the accumulative drifts of radar odometry. One important extension direction of this work is to reduce the prediction outliers, which significantly influence the performance of the large-scale localization. The other direction is to integrate the deep radar relocalization system with deep radar odometry to provide a superior localization system in the real world. In the future, we plan to collect more radar sensory data to supplement the shortage of open dataset, and test our methods on it.

\addtolength{\textheight}{-12cm}   



\section*{ACKNOWLEDGMENT}

This work was supported in part by the NIST grant 70NANB17H185 and UKRI EP/S030832/1 ACE-OPS. The authors would like to thank Dr. Sen Wang for the fruitful discussion and suggestions.

\bibliographystyle{abbrv}
\bibliography{references}

\end{document}